\pdfoutput=1
\documentclass[11pt]{article}

\usepackage{EMNLP2022}

\usepackage{times}
\usepackage{latexsym}
\usepackage{xcolor}

\usepackage[T1]{fontenc}
\usepackage[utf8]{inputenc}

\usepackage{microtype}

\usepackage{inconsolata}

\usepackage{graphicx}

\title{Validity Assessment of Legal Will Statements\\
as Natural Language Inference}

\author{Alice Saebom Kwak, Jacob O. Israelsen, Clayton T. Morrison, \\ {\bf Derek E. Bambauer}, {\bf Mihai Surdeanu} \\
  The University of Arizona, Tucson, Arizona, USA \\
  {\tt\{alicekwak, jisraelsen, claytonm, derekbambauer, msurdeanu\}@arizona.edu}}

\begin{document}
\maketitle
\begin{abstract}
This work introduces a natural language inference (NLI) dataset that focuses on the validity of statements in legal wills. This dataset is unique because: (a) each entailment decision requires three inputs: the statement from the will, the law, and the conditions that hold at the time of the testator's death; and (b) the included texts are longer than the ones in current NLI datasets. We trained eight neural NLI models in this dataset. All the models achieve more than 80\% macro F1 and accuracy, which indicates that neural approaches can handle this task reasonably well. However, group accuracy, a stricter evaluation measure that is calculated with a group of positive and negative examples  generated from the same statement as a unit, is in mid 80s at best, which suggests that the models' understanding of the task remains superficial. Further ablative analyses and explanation experiments indicate that all three text segments are used for prediction, but some decisions rely on semantically irrelevant tokens. This indicates that overfitting on these longer texts likely happens, and that additional research is required for this task to be solved.

\end{abstract}

\section{Introduction}

Natural language inference (NLI) in the legal domain has not been widely investigated, despite its importance and potential. One such important NLI application is validity assessment of legal documents such as wills. Legal procedures for creating and executing wills are evolving rapidly. Processing a will via probate is a costly, time-consuming process that can be exacerbated by errors or by challenges to validity. These problems will likely increase as people increasingly employ electronic wills. Developing natural language techniques that can determine a will’s validity at creation and execution can increase validity, conserve legal resources, and effectuate the author’s intent. To this end, this work explores how NLI models can be employed to evaluate the validity of will statements.
%Alice: a part of the introduction was revised by Guy and Professor Bambauer

NLI in general deals with determining whether a premise entails, contradicts, or is neutral to the hypothesis given. Our project made two important changes to this approach to fit legal documents. First, our dataset contains three input types: statements from wills (as hypotheses), laws (as premises), and {\em conditions}, which are circumstances at the time of probate (e.g., eligibility of the testator or witnesses). This adaptation is necessary as will statements’ validity cannot be evaluated without considering both circumstances at the time of probate (“conditions”) and relevant legal rules (“laws”). Also, we switched the labels from entailment, contradict, neutral to support, refute, unrelated to better represent the relation between will statements and laws. \\
The major contributions of this work are:
\begin{itemize}
    \item We create an open-access annotated dataset with 1,014 data points, generated from 23 publicly available wills. In addition to the three-tuple setting unique to the legal domain, this dataset also contains texts considerably longer than in other open-domain NLI datasets, such as SNLI \cite{snli:emnlp2015}. The average length of our texts is 269 tokens, while the average lengths of premises and hypotheses in SNLI are 8 and 14 tokens, respectively.
    \item We demonstrate that validity assessment of legal will statements can be handled reasonably well with state-of-the-art NLI models when trained on our dataset. However, low scores in group accuracy, which is a stricter evaluation measure calculated with a group of positive and negative examples generated from the same statement as a unit, indicate that work remains before NLI models fully understand legal language.
    \item We explain how the trained models utilize our dataset via ablation tests that remove various pieces of information at prediction time (laws, conditions, or both), and through post-hoc explainability analyses using LIME \cite{ribeiro2016should}. Our analyses indicate that in most cases the NLI classifiers use meaningful phrases from all three pieces of text, indicating the models do capture useful information. However, in some situations, the models use features that are not intuitive for humans, indicating that the task is incomplete.
\end{itemize}
% - Created an annotated dataset with 1,014 data point generated from 23 publicly available wills \\
% - Demonstrated that validity assessment of legal will statements can be handled reasonably well with state-of-the-art NLI models when trained with our dataset \\
% - Explained how the trained models utilizes our dataset by referring to the results from ablation experiment and LIME implementation \\

% These instructions are for authors submitting papers to EMNLP 2022 using \LaTeX. They are not self-contained. All authors must follow the general instructions for *ACL proceedings,\footnote{\url{http://acl-org.github.io/ACLPUB/formatting.html}} as well as guidelines set forth in the EMNLP 2022 call for papers. This document contains additional instructions for the \LaTeX{} style files.
% The templates include the \LaTeX{} source of this document (\texttt{emnlp2022.tex}),
% the \LaTeX{} style file used to format it (\texttt{emnlp2022.sty}),
% an ACL bibliography style (\texttt{acl\_natbib.bst}),
% an example bibliography (\texttt{custom.bib}),
% and the bibliography for the ACL Anthology (\texttt{anthology.bib}).

\section{Related Work}
Natural language inference (NLI), also known as Recognizing Textual Entailment (RTE), determines entailment relations between a pair of sentences: a premise and a hypothesis. Their relationship is either: a) entailment, if a premise entails a hypothesis; b) contradiction, if a premise contradicts a hypothesis; or c) neutral, if a premise neither entails nor contradicts a hypothesis. NLI has been a key framework in natural language processing since \citet{10.1007/11736790_9} proposed the RTE challenge.

Numerous datasets exist for general NLI tasks. For example, the Stanford Natural Language Inference (SNLI) Corpus \cite{snli:emnlp2015} dataset contains about 570,000 pairs of sentences (premises and hypotheses) generated from Flickr image captions.  More recently, datasets for domain-specific NLI tasks are being introduced. SciNLI \cite{sadat-caragea-2022-scinli} is drawn from scientific texts. It contains 107,412 sentence pairs extracted from papers in natural language processing and computational linguistics.

There have been a few NLI-related resources for the legal domain, including IFlyLegal \cite{wang-etal-2019-iflylegal}, StAtutory Reasoning Assessment (SARA) dataset \cite{DBLP:journals/corr/abs-2005-05257}, Graph-based Causal Inference (GCI) framework \cite{DBLP:journals/corr/abs-2104-09420}, and ContractNLI \cite{DBLP:journals/corr/abs-2110-01799}. One IFlyLegal module was developed for natural language article inference; it provides relevant legal articles when asked legal questions. SARA contains rules extracted from the US Internal Revenue Code and natural language questions. GCI generates causal graphs based on fact descriptions. Though the framework was not specifically designed for law, \citet{DBLP:journals/corr/abs-2104-09420} demonstrated that it can be utilized for legal text analysis. ContractNLI is a dataset consisting of 607 legal contracts designed to handle document-level natural language inference.

%The Internal Revenue Code is a mix of statute and administrative regulation, just FYI...

Our work differs from these existing NLI works in two ways: 
% ms: this doesn't seem that important..
%a) rather than determining text entailment relationship, it aims to evaluate the validity of legal will statements with the labels of support, refute, and unrelated, 
(a) it contains three types of input information; and (b) it operates in a pragmatic middle space with respect to text length: larger than datasets with sentence-level texts, which are insufficient to capture legal details, but shorter than document-level inference datasets. 

\graphicspath{ {./images/} }

\begin{figure*}[t!]
  \includegraphics[width=\textwidth]{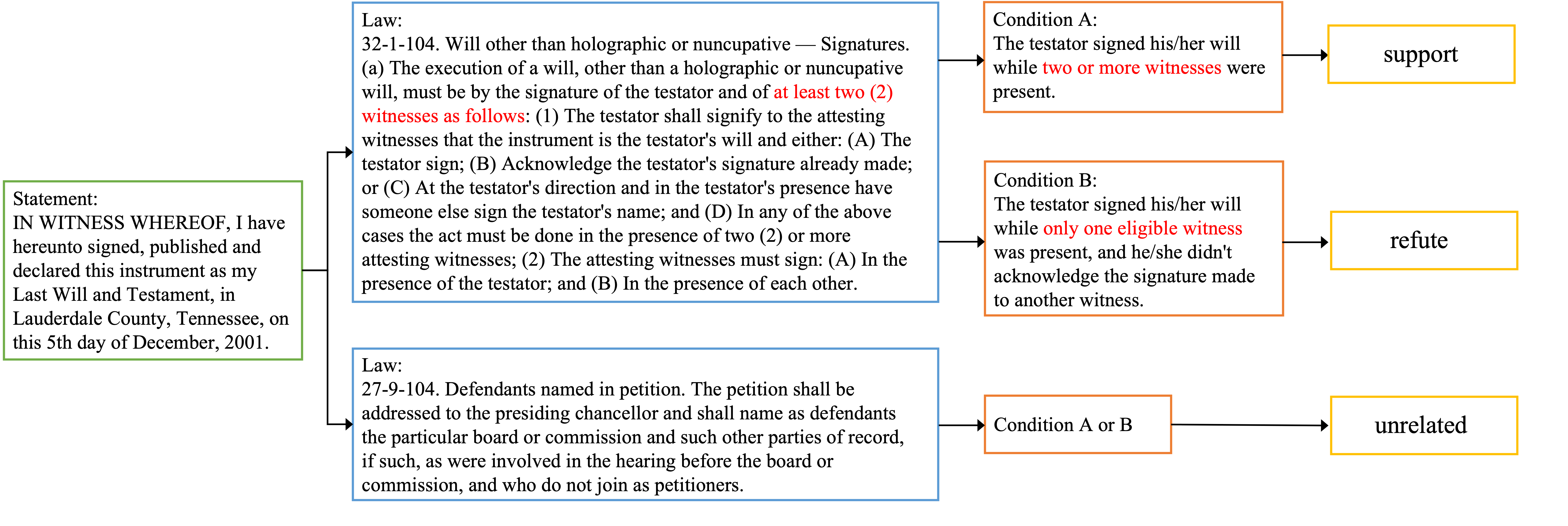}
  \caption{A demonstration of the annotation procedure. Given a law and condition, each statement is evaluated for validity. {\em Support}: it is supported by the law and condition provided; {\em Refute}: refuted by the law and condition. {\em Unrelated}: not relevant to the law. When a statement and a law are unrelated to each other, a condition is not required for classification. However, it is still assigned one of the conditions from either support or refute case. The goal is
  to prevent NLI models from depending on contextual difference (i.e., “texts without conditions are all unrelated
  cases”) rather than features when making predictions.}
\end{figure*}

\section{Dataset}

Our dataset includes three types of entries (will statements, applicable law, and facts representing external state) rather than the usual two types. Validity often depends upon such external circumstances. Thus, a will statement can be legally valid under some facts, but not others. Hence, a will can change from valid to invalid or vice versa; wills must, at minimum, be evaluated when executed (drafted and signed) and when probated (where a court determines whether to implement its provisions). For example, if a will contains a statement appointing a specific person as executor, that person's eligibility must be verified. This requires information about both the law and the person's circumstances at probate time. For example, according to Tennessee Code section 40-20-115, any person who has been imprisoned cannot be an executor. Thus, to assess eligibility of a Tennessee executor in Tennessee, one must know whether they have been in prison at probate time. % ms: nice example

%Derek: this isn't very elegant so I would appreciate feedback.

\subsection{Data Collection}
Our data was collected from the U.S. Wills and Probates datasets in Ancestry, which contains documents in the public domain.\footnote{Court documents are in the public domain in the US.} 
%{Correct Alice or Derek?} Alice: I believe Ancestry counts as a public domain, though it requires membership subscription. Derek: the wills themselves are public documents. It's essentially a bright line rule that court documents are in the public domain. I believe all of the wills we have are ones that have been probated, so that solves the problem of IP claims from the testator's estate or the drafting attorneys. In theory, Ancestry might try to assert some sort of claim against us, but there are two important barriers that protect us: 1) we're operating within their license agreement in my opinion, and 2) even if not, we would almost certainly be shielded by fair use or lack of copyright-eligible subject matter (images of public domain documents probably cannot be copyrighted). TL;DR: public domain is correct.
We chose 23 wills based on three criteria: 1) whether the wills were handwritten; 2) the wills' execution and probation dates; and 3) where the wills were executed and probated. 
We excluded handwritten wills due to the difficulty of OCR text recognition. Execution and probation dates can affect interpretation of wills; we excluded wills executed before 1970 and probated before 2000. Lastly, we only collected wills from Tennessee, because including wills from multiple states would require analyzing each state's laws governing wills. Tennessee had the greatest number of wills meeting our criteria.

All collected wills were anonymized. Personal information was replaced with special tokens denoting the type of information (such as [Person-$n$], [Address-$n$], [Number-$n$], where $n$ identified each person or object in the will. (\citet{suntwal-etal-2019-importance} suggested this anonymization method). Anonymization was performed manually to prevent including personal information in the dataset.

\begin{figure*}
\centering
  \caption*{a) Texts (statement+condition+law)}
%Alice: I think this was mixed up by a mistake! I apologize for the confusion!
  \includegraphics[width=0.5\textwidth]{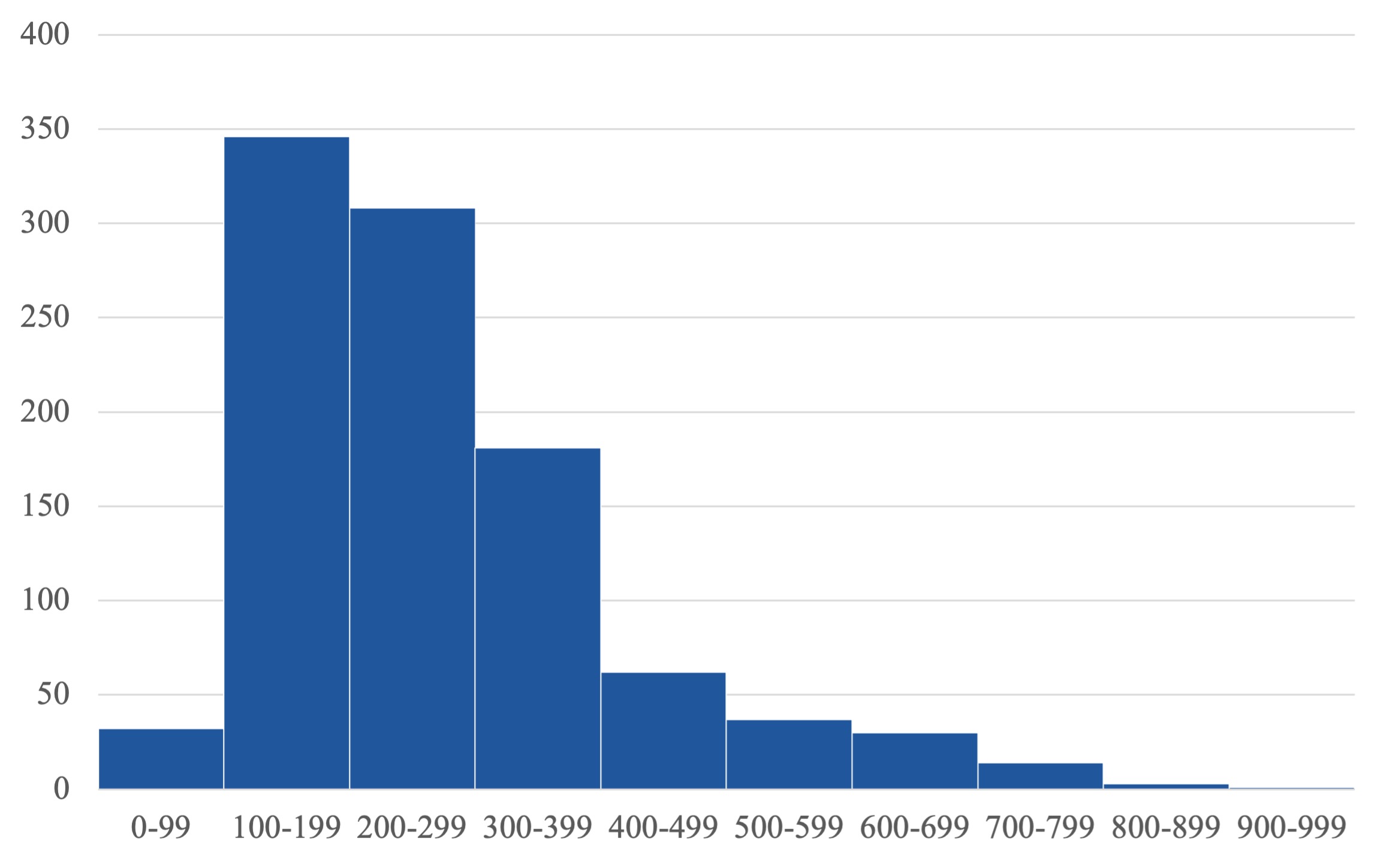}
  \caption*{b) Statements \hspace{3.3cm} c) Conditions \hspace{3.3cm} d) Laws}
  \includegraphics[width=0.32\textwidth]{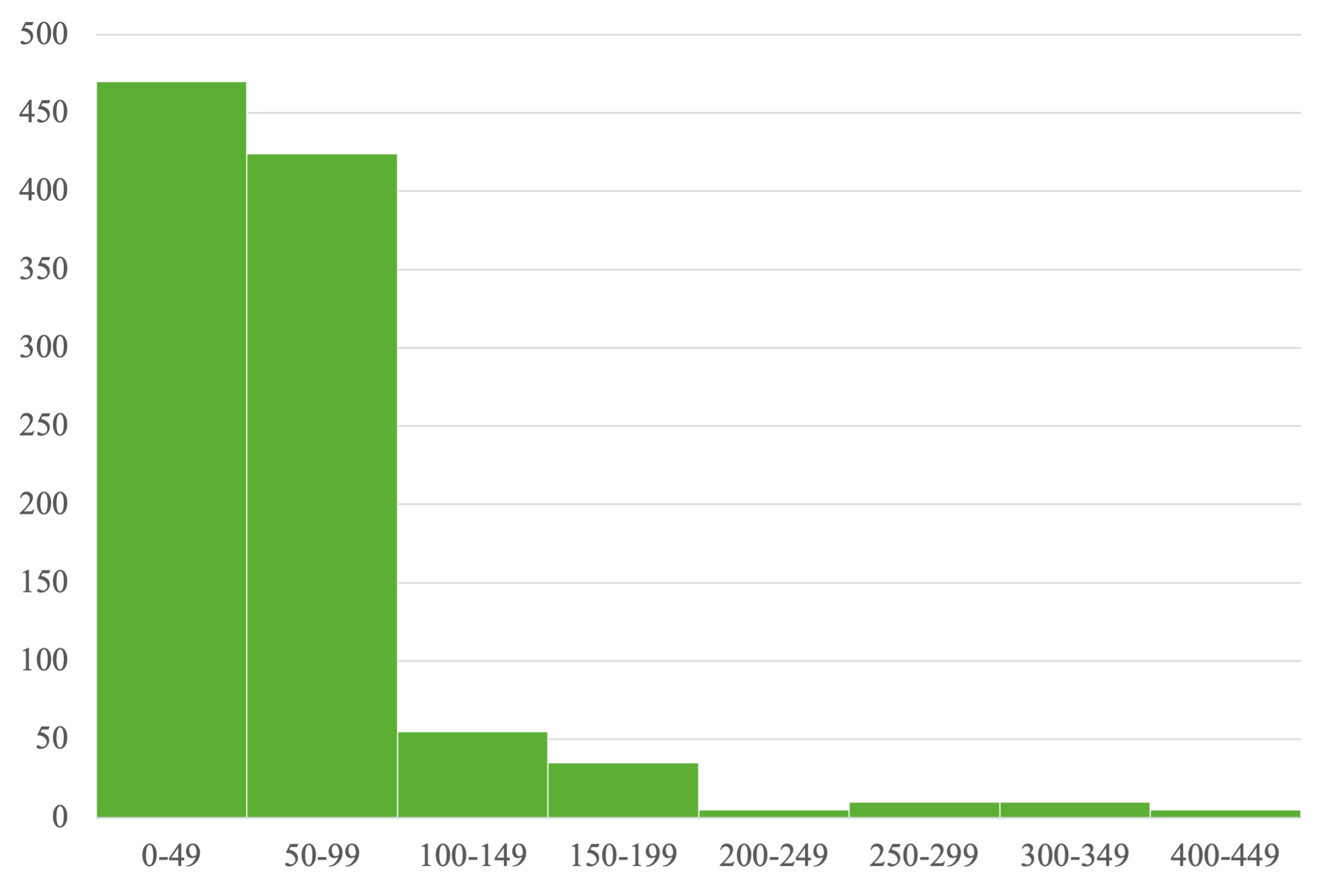}
  \includegraphics[width=0.32\textwidth]{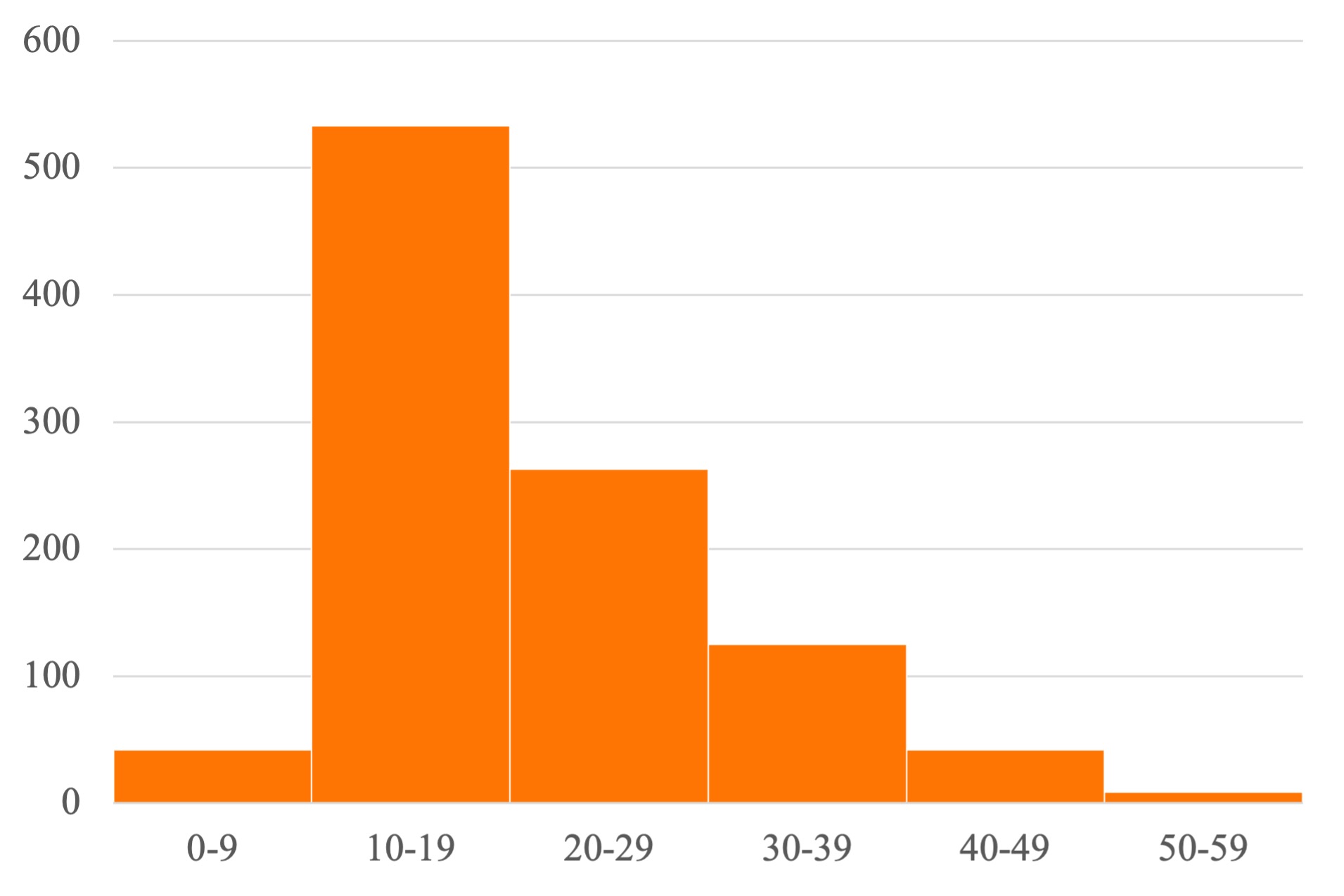}
  \includegraphics[width=0.32\textwidth]{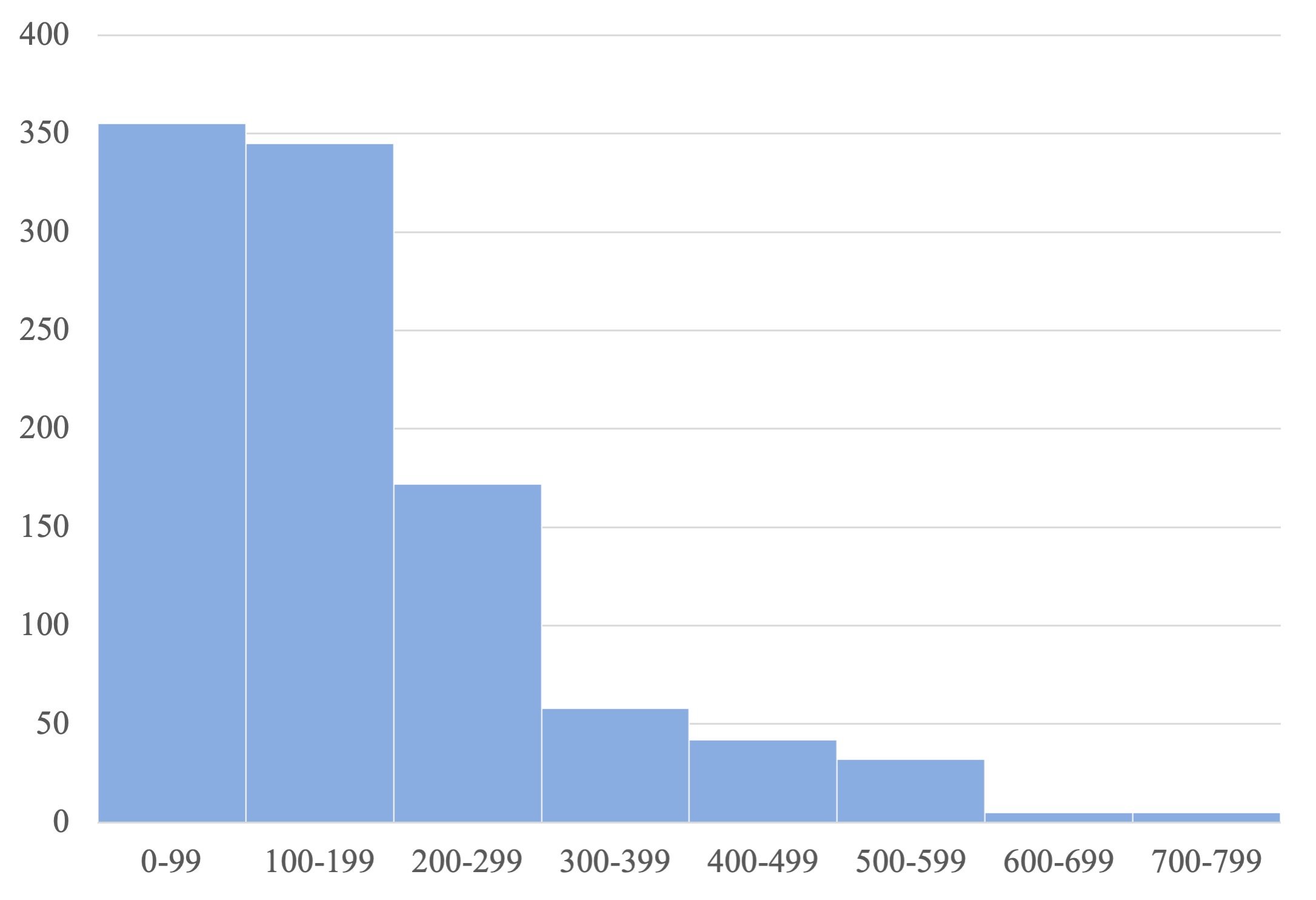}
% \end{figure*}
% \begin{figure*}
%   \includegraphics[width=0.3\textwidth]{histogram_condition}
% \end{figure*}
% \begin{figure*}
%   \includegraphics[width=0.3\textwidth]{histogram_law}
  \caption{Histograms demonstrating the length distribution of a) full texts (i.e., statement+condition+law), b) statements, c) conditions, and d) laws in terms of token counts. Token counts are plotted on the X-axis; the Y-axis indicates the number of texts/statements/conditions/laws for each bin.}
  \vspace{-4mm}
\end{figure*}

% \begin{figure*}
% \centering
%   \caption*{a) Texts (statement+condition+law)}
% %Alice: I think this was mixed up by a mistake! I apologize for the confusion!
%   \includegraphics[width=0.5\textwidth]{Histogram_total_retruncated.jpg}
%   \caption*{b) Statements \hspace{3.3cm} c) Conditions \hspace{3.3cm} d) Laws}
%   \includegraphics[width=0.32\textwidth]{Histogram_statements_retruncated.jpg}
%   \includegraphics[width=0.32\textwidth]{Histogram_conditions_retruncated.jpg}
%   \includegraphics[width=0.32\textwidth]{Histogram_laws_retruncated.jpg}
% % \end{figure*}
% % \begin{figure*}
% %   \includegraphics[width=0.3\textwidth]{histogram_condition}
% % \end{figure*}
% % \begin{figure*}
% %   \includegraphics[width=0.3\textwidth]{histogram_law}
%   \caption{Histograms demonstrating the length distribution of a) full texts (i.e., statement+condition+law), b) statements, c) conditions, and d) laws in terms of token counts. Token counts are plotted on the X-axis; the Y-axis indicates the number of texts/statements/conditions/laws for each bin.}
%   \vspace{-4mm}
% \end{figure*}

\subsection{Annotation}
Each will statement was annotated as {\em support}, {\em refute}, or {\em unrelated} based on a given condition and law. (Our annotators set hypothetical conditions when necessary to categorize statements.)

% Alice: I added a sentence in the paragraph below, to provide more accurate explanation on the annotation process. Please let me know if it doesn't make sense.
% Derek: I would explain a bit more: we included [IIRC] a condition making it true, a condition making it false, and also three different types of laws. I'm having a bit of trouble following the procedure here.
Importantly, each statement was annotated multiple times. A statement was annotated once as support, once as refute, and thrice as unrelated.\footnote{A few exceptions occurred when: \\
a) it was impossible to create a case where the statement was refuted. In such cases, statements were not annotated as "refute." \\
b) a statement had more than one relevant legal rule. Such statements were annotated twice as support, twice as refute, and once as unrelated.} This ensured that every statement was labeled with all three classes in the same ratio (support:refute:unrelated=1:1:3). Also, annotating statements three times more with “unrelated” than support and refute enabled including a greater range of laws in the dataset.
Formally, our annotation procedure included the following steps: 

{\flushleft {\bf (1)}} preprocessing: extracting texts from collected wills using OCR\footnote{The collected wills were scanned documents, so OCR was necessary to extract machine-readable text.}, and copying statements into the dataset;
{\flushleft {\bf (2)}} for each statement, identifying the Tennessee legal provision that supported, refuted, or was unrelated to the statement's validity;
{\flushleft {\bf (3)}} adding a condition specifying external circumstances relevant to validity;
{\flushleft {\bf (4)}} repeating these steps to generate five annotations per statement;
{\flushleft {\bf (5)}} anonymizing all statements by replacing personal information with special tokens.

Figure 1 demonstrates the annotation procedure with examples.

Two annotators participated in the task. One is a law student, and annotation was supervised by author Bambauer, a law professor. The other annotator does not have legal training, but ongoing discussions and reviews ensured the uniform quality. The annotators contributed equally. Annotators worked individually but shared annotation guidelines. After the annotation was complete, cross-annotation was conducted on 200 randomly selected items (100 items from each annotator's set; each annotator worked blinded on items drawn from the other annotator's set) to check inter-annotator agreement. We calculated Kappa agreement score based on the cross-annotation result, and the score is 0.91 (rounded to hundredth).

%Derek: dumb question - if the annotators collaborated, doesn't that undercut our Kappa analysis? After all, we'd largely expect consensus, no?

After the annotation was complete, the dataset was split into training, development, and test sets. There are 1014 data points in our dataset, split 50:25:25\% (train: 504, development: 255, test: 255). When splitting, a group containing annotations from a single statement (one support, one refute, and three unrelated) was treated as a single unit. Each annotator's work was equally represented in all three datasets. All 23 wills appear in more than one of the train, dev, and test partitions. However, given the nature of the will statements from the wills and how they were annotated, data leakage is unlikely. Will statements are independent and generally do not convey information about other statements from the same wills. Summary statistics are in Table 1 and Figure 2. % Alice: I believe this was my mistake! It's fixed now.

\begin{table}
\centering
\begin{tabular}{lll}
\hline
\textbf{Datasets} & \textbf{Texts} & \textbf{Tokens}\\
\hline
Train & 504 & 135218 \\
Development & 255 & 69643 \\
Test & 255 & 68042 \\ 
\hline
\textbf{Total} & \textbf{1014} & \textbf{272903} \\ 
\hline
\end{tabular}
\caption{The number of texts and tokens contained in each dataset and in total}
\label{tab:accents}
\end{table}

\subsection{Characteristics of the Dataset}
Our dataset has two characteristics distinguishing it from other datasets: a) texts are composed of three input types, and b) texts contain a large number of tokens. 

First, our task requires a third input type: a condition, since validity often depends upon conditions.

Second, our texts tend to contain a large number of tokens. 44\% of our texts contain 200 or more tokens. The average token number was 269.14; 79 texts contain more than 512 tokens (the threshold for most NLI models). Popular NLI datasets are shorter.

\subsection{Implementation}
We trained multiple NLI models with our dataset to assess their performance: four transformer models and four sentence-transformer models. The transformer models include bert-base-uncased \cite{DBLP:journals/corr/abs-1810-04805}, distilbert-base-uncased \cite{DBLP:journals/corr/abs-1910-01108}, roberta-large-mnli \cite{liu2019roberta}, and longformer-base-4096 \cite{Beltagy2020Longformer}. Bert-base-uncased and dilstilbert-base-uncased were trained to set baselines on the task. In addition to the baseline models, we used roberta-large-mnli and longformer-base-4096. Among sentence-transformer models \cite{reimers-2019-sentence-bert}, four pretrained models with top average performances (based on the Model Overview page\footnote{\url{https://www.sbert.net}}) were chosen: all-mpnet-base-v2\footnote{\url{https://huggingface.co/sentence-transformers/all-mpnet-base-v2}}, multi-qa-mpnet-base-dot-v1\footnote{\url{https://huggingface.co/sentence-transformers/multi-qa-mpnet-base-dot-v1}}, all-distilroberta-v1 \cite{Sanh2019DistilBERTAD}\footnote{\url{https://huggingface.co/sentence-transformers/all-distilroberta-v1}}, and all-MiniLM-L12-v2 \cite{wang2020minilm}\footnote{\url{https://huggingface.co/sentence-transformers/all-MiniLM-L12-v2}}.

To distinguish between statements, laws, and conditions in the concatenated texts, we prefixed each with a special token: {\tt [STATE]}, {\tt [LAW]}, and {\tt [COND]}.

Both transformer models and sentence-transformer models were trained on PyTorch 1.11.0 with Cuda 11.3. All transformer models were trained using the Trainer class provided by HuggingFace. Learning rates and training epochs were tuned on the development partition. Sentence-transformer models were trained with Sentence Transformer Fine-Tuning (SetFit) proposed by \citet{wasserblat_2021}. It was slightly adapted to fit the multi-class classification task, but its fundamentals remain intact. SetFit utilizes sentence pairs generated within the same class as training data. It fits the model with the data to minimize the Softmax Loss. Once the model is fitted, it is used to encode the training and development (or test, when it is a testing phase) datasets. The encoded data is used to fit the Logistic Regression classifier (with ‘liblinear’ solver) which, finally, makes predictions.

All models except for longformer-base-4096 were trained and evaluated on the truncated datasets due to the models' token number limitations. The models can only process texts with 512 or fewer tokens without truncation (all-MiniLM-L12-v2: 256 tokens; all-mpnet-base-v2: 384 tokens; other models, except longformer-base-4096: 512 tokens). When truncating datasets, the ratio of average token numbers among three input types (statements, laws, and conditions) was considered to prevent losing excessive information from a single type of input.

%Derek: just an explanatory note--"i.e.," is used as shorthand for "That is," meaning that the parenthetical restates the information in a different way to improve comprehension. "e.g.," means "for example"; the parenthetical provides an example of the concept just described, and assumes that the reader understands the concept.

\section{Results and Analysis}

\begin{table*}[ht!]
\centering
{\begin{tabular}{llllll}
\hline
\textbf{Model} & \textbf{Precision} & \textbf{Recall} & \textbf{F1} & \textbf{Accuracy} & \textbf{GA} \\
\hline Transformers models \\
\hline
bert-base-uncased & 81.47 & 82.20 & 81.81 & 87.06 & 49.02\\
distilbert-base-uncased & 82.37 & 82.59 & 82.48 & 87.06 & 50.98\\
longformer-base-4096 & 94.09 & 93.69 & 93.85 & 94.90 & 74.51 \\
roberta-large-mnli & \textbf{96.67} & \textbf{96.25} & \textbf{96.42} & \textbf{96.86} & \textbf{84.31}
\\
\hline Sentence-Transformers models \\
\hline
all-MiniLM-L12-v2 & 81.50 & 84.54 & 82.64 & 85.49 & 50.98 \\
all-disilroberta-v1 & 83.43 & 85.99 & 84.22 & 87.06 & 49.02\\
multi-qa-mpnet-base-dot-v1 & 91.48 & 94.13 & 92.65 & 93.73 & 74.51\\
all-mpnet-base-v2 & 92.93 & 95.05 & 93.82 & 94.90 & 76.47\\
\hline
\end{tabular}}
\caption{\label{results from the trained data}
Results of all classifiers trained on our dataset using five measures: precision, recall, F1, accuracy, and group accuracy (GA in table, see Section~\ref{sec:measures}). Precision, recall, and F1 scores were computed with Macro average. Group accuracy is calculated with positive and negative examples generated from the same statement as a unit. If a group contains one or more incorrect predictions, the group is considered incorrect.
}
\end{table*}

\subsection{Evaluation Measures}
\label{sec:measures}

We report results using standard accuracy, precision, recall, and F1 scores. To ensure all labels are well represented, precision, recall, and F1 scores were computed with Macro average.
Additionally, as suggested by \citet{elazar-etal-2021-back}, we introduce a new measure called 
group accuracy (GA).  GA is calculated with a group of positive and negative examples from the same statement (rather than each text) as a unit. If a group has one or more incorrect predictions, the group is incorrect. 
%This measure is included to evaluate the models’ understanding of will statements on a stricter level. % ms: somewhat content free
If a model correctly understands a will statement, it should perform equally well on all examples from the same statement.

\begin{table*}[ht!]
\centering
% \begin{small}
\begin{tabular}{lllll}
\hline
\textbf{Model} & \textbf{Precision} & \textbf{Recall} & \textbf{F1} & \textbf{Accuracy} \\
\hline all-mpnet-base-v2 (sample n = 52)\\
\hline
token <= 192 & 93.75 & 93.15 & 92.91 & 94.23 \\
192 < token <= 384 & 90.48 & 89.10 & 89.63 & 94.23 \\
384 < token & 86.35 & 89.49 & 87.08 & 88.46 \\
\hline longformer-base-4096 (sample n = 20)\\
\hline
token <= 256 & 95.24 & 88.89 & 90.77 & 95.00 \\
256 < token <= 512 & 80.56 & 71.67 & 70.93 & 80.00 \\
512 < token & 73.81 & 66.67 & 66.93 & 75.00 \\
\hline roberta-large-mnli (sample n = 20)\\
\hline
token <= 256 & 100 & 100 & 100 & 100 \\
256 < token <= 512 & 94.87 & 85.00 & 88.76 & 90.00 \\
512 < token & 94.44 & 93.33 & 93.27 & 95.00 \\
\hline
\end{tabular}
% \end{small}
\label{results with different input lengths}
\caption{Results with different input lengths. Models generally performed worse when token numbers in inputs were larger. Roberta-large-mnli performed better than longformer-base-4096 in predicting long inputs (token > 512) despite truncation. Roberta-large-mnli also showed better performance on long inputs than regular inputs, indicating input truncation did not worsen the model's performance.
}
\end{table*}

\subsection{Results from Trained Models}
Table 2 shows the models’ performances on our test partition.
Overall, the models demonstrated good performances. Each achieved more than 80\% in all metrics but group accuracy. Roberta-large-mnli showed the best performance. It achieved over 96\% in all metrics but group accuracy (84.31\%), suggesting it can handle the task reasonably well. However, the large difference between accuracy and group accuracy suggests that the models' understanding of the task is superficial. 

Accuracy for the {\em unrelated} label was higher than for {\em support} and {\em refute} in all models except all-mpnet-base-v2 and multi-qa-mpnet-base-dot-v1. This higher accuracy for {\em unrelated} label may be partially attributable to the gap between accuracy and group accuracy, as it would inflate overall accuracy. However, since the gap between accuracy and group accuracy is not significantly smaller for all-mpnet-base-v2 and multi-qa-mpnet-base-dot-v1 (i.e., the models where accuracy for {\em unrelated} was not higher than the other labels), it is more likely that the gap originated from the strictness of group accuracy and the models' superficial understanding of will statements.

Further, roberta-large-mnli (trained with truncated inputs) showed better performance than longformer-base-4096 (trained with full length inputs). This suggests that models with token number limitations can still perform well with long inputs when truncated properly.

\begin{table*}[ht!]
\centering
% \begin{small}
\begin{tabular}{llllll}
\hline
\textbf{Model} & \textbf{Precision} & \textbf{Recall} & \textbf{F1} & \textbf{Accuracy} & \textbf{GA} \\
\hline Statements and laws \\
\hline
all-mpnet-base-v2 & 75.64 & 76.47 & 75.98 & 81.96 & 27.45 \\
longformer-base-4096 & 75.72 & 75.98 & 75.79 & 82.75 & 37.25 \\
roberta-large-mnli & 77.93 & 77.51 & 77.55 & 84.31 & 39.22 \\
\hline Statements and conditions \\
\hline
all-mpnet-base-v2 & 46.67 & 57.13 & 44.80 & 44.31 & 0.00 \\
longformer-base-4096 & 58.00 & 49.44 & 51.30 & 58.82 & 0.00 \\
roberta-large-mnli & 37.63 & 41.72 & 38.08 & 56.86 & 0.00 \\
\hline Statements only \\
\hline
all-mpnet-base-v2 & 36.76 & 35.94 & 34.04 & 36.47 & 0.00\\
longformer-base-4096 & 18.17 & 33.33 & 23.52 & 54.51 & 0.00 \\
roberta-large-mnli & 18.17 & 33.33 & 23.52 & 54.51 & 0.00 \\
\hline
\end{tabular}
% \end{small}
\label{results from the ablation experiment}
\caption{Results from the ablation experiment where three models were trained with datasets lacking one (law or condition) or two types (law+condition) of inputs. Results demonstrate that the models' performances deteriorate if one or more input type(s) is omitted. Thus, both types of inputs affect the models' performances, though laws have larger impact than conditions.
}
\end{table*}

\subsection{Experiment with Different Input Lengths}
To investigate whether the models' performance was affected by input length and/or input truncation, we varied input lengths. Inputs were classified into three categories by length: a) short (equal or less than 192 or 256 tokens); b) regular (larger than 192 or 256 tokens and equal or less than 384 or 512); and c) long (larger than 384 or 512)\footnote{For all-mpnet-base-v2 model, 192 and 384 were used as dividing points instead of 256 and 512, as its maximum token length is 384.}. The number of inputs in each category varied. To control the impact of varying sample size, categories with larger sample sizes were reduced by random sampling to match the smallest sample size. For this experiment, we used the three models with best performance on the full dataset.

Table 3 shows the results. Models generally performed worse with larger input token numbers. All three models performed best with inputs with smaller token numbers. One interesting finding is that roberta-large-mnli performed better than longformer-base-4096 in predicting inputs with large token numbers (n > 512) despite truncation. Roberta-large-mnli showed better performance on long inputs than regular inputs, indicating truncation did not negatively affect the model's performance. This finding aligns with the observation from the overall results that models with token number limitation can still show good performances on long inputs when properly truncated. However, results differed with a smaller token number limitation. All-mpnet-base-v2 performed worst on long inputs (where truncation occurred), indicating truncation negatively impacted performance.

\subsection{Results from Ablation Experiment}
To determine whether the models correctly rely on all three types of texts (statements, conditions, and laws), we conducted an ablation experiment, training the models with datasets lacking one (law or condition) or two types (law+condition) of inputs. \citet{poliak-etal-2018-hypothesis} suggested a similar experiment to this one to set hypothesis-only baselines. We used the three models with the best performances on the full dataset for this experiment.

Table 4 shows the results. Performance deteriorated if any input type was omitted. Among the models trained on partial data, those trained with statements and laws performed better, with F1 scores over 75\%. However, group accuracy dropped to the 30s or even 20\%s, indicating actual understanding decreased considerably.
Models trained with statements and conditions performed significantly worse. Results ranged between 38\% and 52\% for F1, and group accuracy dropped to 0. Models trained only with statements performed worst, with F1 scores dropping to 35\% or lower. 
% the results ranging from 18\% to 55\% except for group accuracy. The difference between these results and the ones from the models trained with statements & conditions is not as significant as the one shown by the former comparison. Still, it is clear that the models trained with both statements and conditions outperforms the models trained only with statements.

This degradation from partial data shows including both conditions and laws positively affects the models’ performance. Models use all three types of inputs when making predictions\footnote{A reviewer suggested an experiment where laws and conditions are packed into premises and treat the task as normal NLI. A pilot experiment with the best performing model (roberta-large-mnli) found a slight deterioration in performance when laws and conditions were combined as premises (precision: 93.09, recall: 93.89, F1: 93.41, accuracy: 94.12, GA: 74.51; see Table 2 for comparison with main result). The biggest difference came from GA score (74.51 vs. 84.31). This lowered GA score implies that the model's understanding worsened when conditions and laws were combined into premises. Given this preliminary result, it is likely that three-way distinction in input types positively contributes to the model's performance. Nevertheless, as the performance difference is not significantly large, further investigation is needed.}. Laws have more impact on performance than conditions, since deterioration in results without laws was more significant than from models trained without conditions. This is realistic, since laws set the parameters within which conditions operate to make a given provision valid or invalid.

% \begin{figure*}[t!]
% \centering
%   \includegraphics[width=1\textwidth]{EMNLP 2022/images/lime_correct1.jpg}
%   \vspace{-10mm}
%   \caption*{a) A text correctly predicted as {\em support}}
%   \includegraphics[width=1\textwidth]{EMNLP 2022/images/lime_correct2.jpg}
%   \vspace{-10mm}
%   \caption*{b) A text correctly predicted as {\em refute}}
%   \vspace{-2mm}
%   \caption{Examples of LIME explanations showing cases where the model (roberta-large-mnli) makes correct predictions based on features sensible to humans.}
%   \vspace{-2mm}
% \end{figure*}

\begin{figure*}[t!]
\centering
  \includegraphics[width=1\textwidth]{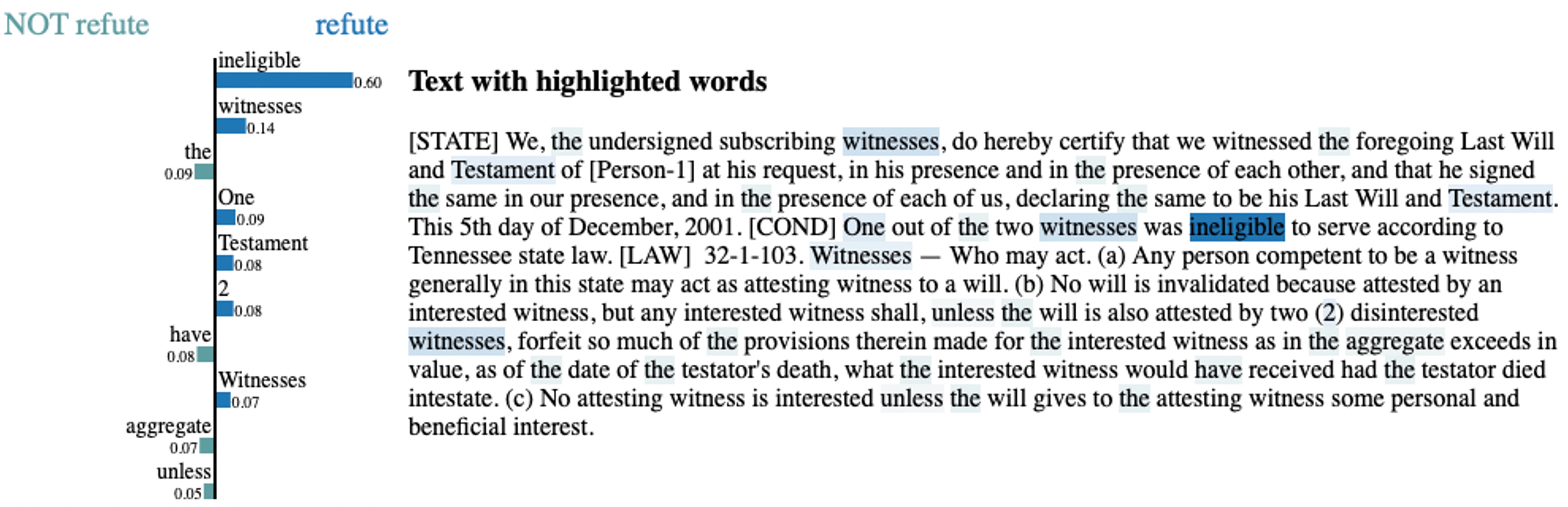}
  \vspace{-2mm}
  \caption{Examples of LIME explanations showing a case where the model (roberta-large-mnli) makes correct predictions based on features sensible to humans.}
  \vspace{-2mm}
\end{figure*}

\begin{figure*}[t]
\centering
  \includegraphics[width=1\textwidth]{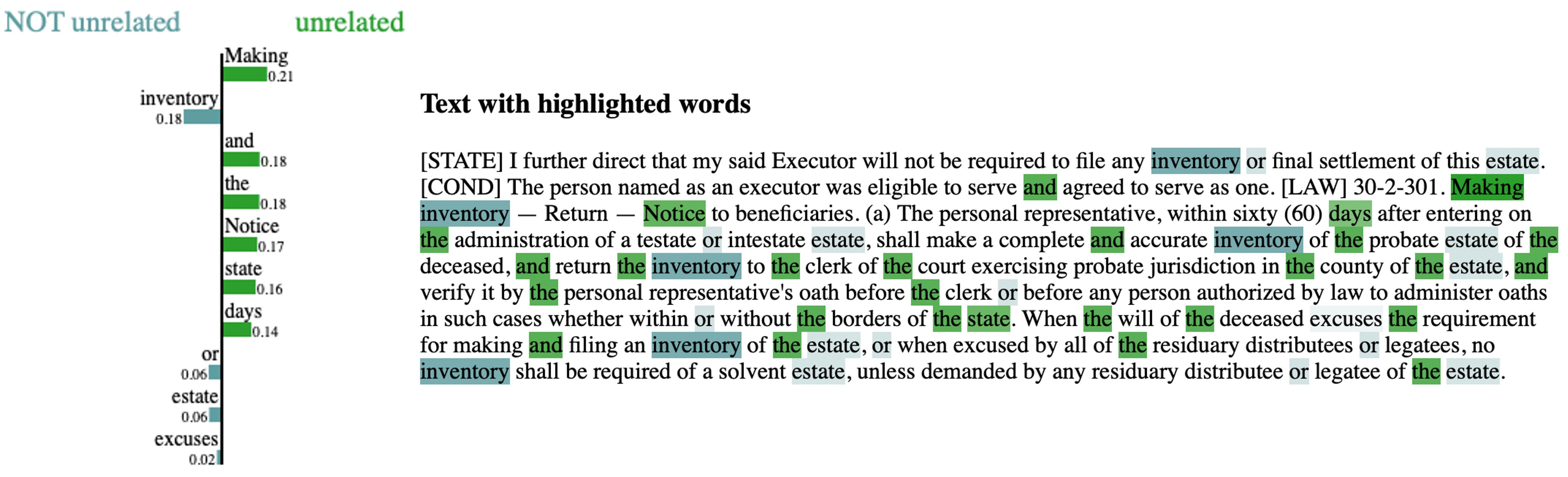}
  \vspace{-4mm}
  \caption{An example of a LIME explanation showing a case where the model (roberta-large-mnli) makes an incorrect prediction based on the tokens such as \textit{Making}, \textit{and}, and \textit{the} which bear little semantic relevancy to the gist of the text.}
\end{figure*}

\subsection{Understanding Results with LIME}
To clarify model behavior, we used 
Local Interpretable Model-Agnostic Explanations (LIME) to explain our classifiers' predictions \cite{ribeiro2016should}. 
%It provides human interpretable explanations on a local level by approximating the local behavior of the classifier with an interpretable model. % ms: well known in the field
We implemented LIME on predictions made by the best performing model (roberta-large-mnli, fine-tuned on our data). The LIME results revealed that the model tends to correctly rely on all three texts (statements, laws, conditions) for a prediction, and uses meaningful features in many cases. However, sometimes the model utilized features that are not intuitive for humans.

\subsubsection{A Correct Example}
Figure 3 shows a case where the model made a correct prediction based on features sensible to humans. Figure 3's text is a statement saying two or more witnesses witnessed the testator signing the will and signed in each other's presence. It includes a condition stating that one of the two witnesses was ineligible to serve under Tennessee law, and the law specifying witness eligibility in Tennessee. The given condition and the law invalidate the will statement, rendering it as {\em refute}. The model correctly predicted the outcome based on tokens such as \textit{ineligible}, \textit{witnesses}, \textit{One}, \textit{testament}, and \textit{2}. The top two tokens with greatest impact on the model’s prediction were \textit{ineligible} and \textit{witnesses}, with importance score of 0.60 and 0.14, respectively. It is understandable that these tokens have high scores, as they are the keywords ("\textbf{One} out of two (\textbf{2}) \textbf{witnesses} is \textbf{ineligible}") which provide grounds for revoking a will statement.

\subsubsection{An Incorrect Example}
Figure 4 shows a case where the model made an incorrect prediction based on irrelevant tokens. The intent of the will statement is to excuse the Executor from filing inventory. This is not contradicted by the condition and law, requiring the model to classify the text as {\em support}. Nevertheless, the model incorrectly classified it as {\em unrelated}. The reason for the incorrect prediction is that the model (probably) relied on irrelevant tokens such as \textit{Making}, \textit{and}, and \textit{the}, which bear little semantic relevancy to the text, versus more relevant tokens such as \textit{inventories} and \textit{excuses} when making the prediction. This is likely due to overfitting induced by longer texts. This LIME explanation shows that even the best performing model still has room for improvement.

\section{Future Work}
This work can be expanded in several directions. First, it can be extended to cover multiple states by adapting the models or adding more data. Future work can investigate if models trained on a single state's data can be adapted to evaluate data from other states. Also, wills from multiple states could be added to the dataset. We expect the augmented dataset would enhance the models' performance on evaluating wills from other states.

This work can be expanded to other legal domains. Models trained on our data can be adapted to similar tasks in other legal domains.

Lastly, this work can be extended by investigating novel transformer models. Given the uniqueness of our dataset with regard to the number of inputs and text length, it is likely that further experiment and modification is needed to handle such characteristics.

\section{Conclusion}
This work presented an annotated dataset for natural language inference (NLI) in the legal domain, consisting of 1,014 data points generated from 23 publicly available wills. The dataset is novel for two reasons: it included texts with three input types (statement, law, and condition) rather than two (premise and hypothesis) in the traditional NLI, and included texts are longer than in general NLI datasets. The NLI models trained on our dataset showed reasonable performance in assessing the validity of will statements. Ablative experiments demonstrated that the models’ performances worsen if any input type (condition, law, or both) is omitted. This suggests that the models utilize all three input types. The LIME implementation reveals that even the best performing model makes errors in some cases by using semantically irrelevant tokens. Our open-access dataset is publicly available at: \url{https://github.com/ml4ai/nli4wills-corpus}

\section*{Limitations}
Our dataset consists of a relatively small number of data points (1,014 texts). Annotating will statements with relevant laws and conditions is a highly demanding, time intensive task. 
% ms: not really supported by GA...
%Considering the relatively high performances of the models trained on our data, we believe that our dataset contains a sufficient amount of information to handle the task. However, 
A larger size of dataset would likely further improve the models’ performances.

Our dataset only includes wills executed and probated in Tennessee, with execution after 1970 and probate after 2000. Due to these restrictions, our framework might produce incorrect results on inputs from wills from different settings. Supplementing the dataset with wills from more diverse settings would address this limitation. Even though the scope is limited to a single state (Tennessee), our study demonstrates that transformer models trained on the dataset can evaluate the validity of statements from wills with reasonable accuracy.
% Alice: Does this addition sound okay? I'm a bit concerned that this sentence doesn't fit well in limitations section.
Future additions to our dataset will be available at the same URL: \url{https://github.com/ml4ai/nli4wills-corpus}

This work does not involve humans in the loop. Considering how crucial accuracy is for the task (i.e., legal validity evaluation), the work would have benefited much from involving domain experts. Even though our study discovered that the state-of-the-art transformer models can show good performances (over 85\% accuracy in all 8 models) without human interaction, it was also found that the models' understanding on the task is rather superficial (GAs ranging from 49-85\% in Table 2). Including humans-in-the-loop could be a solution for enhancing the models' understanding on the task.

\section*{Ethics Statement}
We collected legal wills from Ancestry as a part of our dataset creation process. The wills probated in court are in the public domain in the US, and we did not violate Ancestry’s Terms and Conditions when collecting wills. We also anonymized the wills by replacing any personal identifiable information contained in the documents with special tokens.

Our datasets and codes are released to public. We believe our released datasets and codes would contribute to society by promoting further NLI endeavors in the legal domain. The resources could potentially assist with people reviewing wills, but they should not be considered as legal advice. To avoid any confusion, we placed a disclaimer that the users must not rely on any information provided from our resources when making legal decisions and should instead consult with an attorney.

%Alice: Ancestry's terms and conditions link -  https://www.ancestry.com/c/legal/termsandconditions
% @professor Bambauer, would you be able to review the statement and give me any feedback?

% ms: we can't include Acks due to the blind submission
\section*{Acknowledgements}
We thank the reviewers for their thoughtful comments and suggestions. This work was partially supported by the National Science Foundation (NSF) under grant \#2217215, and by University of Arizona's Provost Investment Fund. Mihai Surdeanu and Clayton Morrison declare a financial interest in lum.ai. This interest has been properly disclosed to the University of Arizona Institutional Review Committee and is managed in accordance with its conflict of interest policies.

\bibliography{anthology,custom}

\begin{thebibliography}{19}
\expandafter\ifx\csname natexlab\endcsname\relax\def\natexlab#1{#1}\fi

\bibitem[{Beltagy et~al.(2020)Beltagy, Peters, and
  Cohan}]{Beltagy2020Longformer}
Iz~Beltagy, Matthew~E. Peters, and Arman Cohan. 2020.
\newblock Longformer: The long-document transformer.
\newblock \emph{arXiv:2004.05150}.

\bibitem[{Bowman et~al.(2015)Bowman, Angeli, Potts, and
  Manning}]{snli:emnlp2015}
Samuel~R. Bowman, Gabor Angeli, Christopher Potts, and Christopher~D. Manning.
  2015.
\newblock A large annotated corpus for learning natural language inference.
\newblock In \emph{Proceedings of the 2015 Conference on Empirical Methods in
  Natural Language Processing (EMNLP)}. Association for Computational
  Linguistics.

\bibitem[{Dagan et~al.(2006)Dagan, Glickman, and Magnini}]{10.1007/11736790_9}
Ido Dagan, Oren Glickman, and Bernardo Magnini. 2006.
\newblock The pascal recognising textual entailment challenge.
\newblock In \emph{Machine Learning Challenges. Evaluating Predictive
  Uncertainty, Visual Object Classification, and Recognising Tectual
  Entailment}, pages 177--190, Berlin, Heidelberg. Springer Berlin Heidelberg.

\bibitem[{Devlin et~al.(2018)Devlin, Chang, Lee, and
  Toutanova}]{DBLP:journals/corr/abs-1810-04805}
Jacob Devlin, Ming{-}Wei Chang, Kenton Lee, and Kristina Toutanova. 2018.
\newblock \href {http://arxiv.org/abs/1810.04805} {{BERT:} pre-training of deep
  bidirectional transformers for language understanding}.
\newblock \emph{CoRR}, abs/1810.04805.

\bibitem[{Elazar et~al.(2021)Elazar, Zhang, Goldberg, and
  Roth}]{elazar-etal-2021-back}
Yanai Elazar, Hongming Zhang, Yoav Goldberg, and Dan Roth. 2021.
\newblock \href {https://doi.org/10.18653/v1/2021.emnlp-main.819} {Back to
  square one: Artifact detection, training and commonsense disentanglement in
  the {W}inograd schema}.
\newblock In \emph{Proceedings of the 2021 Conference on Empirical Methods in
  Natural Language Processing}, pages 10486--10500, Online and Punta Cana,
  Dominican Republic. Association for Computational Linguistics.

\bibitem[{Holzenberger et~al.(2020)Holzenberger, Blair{-}Stanek, and
  Durme}]{DBLP:journals/corr/abs-2005-05257}
Nils Holzenberger, Andrew Blair{-}Stanek, and Benjamin~Van Durme. 2020.
\newblock \href {http://arxiv.org/abs/2005.05257} {A dataset for statutory
  reasoning in tax law entailment and question answering}.
\newblock \emph{CoRR}, abs/2005.05257.

\bibitem[{Koreeda and Manning(2021)}]{DBLP:journals/corr/abs-2110-01799}
Yuta Koreeda and Christopher~D. Manning. 2021.
\newblock \href {http://arxiv.org/abs/2110.01799} {Contractnli: {A} dataset for
  document-level natural language inference for contracts}.
\newblock \emph{CoRR}, abs/2110.01799.

\bibitem[{Liu et~al.(2021)Liu, Yin, Feng, Wu, and
  Zhao}]{DBLP:journals/corr/abs-2104-09420}
Xiao Liu, Da~Yin, Yansong Feng, Yuting Wu, and Dongyan Zhao. 2021.
\newblock \href {http://arxiv.org/abs/2104.09420} {Everything has a cause:
  Leveraging causal inference in legal text analysis}.
\newblock \emph{CoRR}, abs/2104.09420.

\bibitem[{Liu et~al.(2019)Liu, Ott, Goyal, Du, Joshi, Chen, Levy, Lewis,
  Zettlemoyer, and Stoyanov}]{liu2019roberta}
Yinhan Liu, Myle Ott, Naman Goyal, Jingfei Du, Mandar Joshi, Danqi Chen, Omer
  Levy, Mike Lewis, Luke Zettlemoyer, and Veselin Stoyanov. 2019.
\newblock Roberta: A robustly optimized bert pretraining approach.
\newblock \emph{arXiv preprint arXiv:1907.11692}.

\bibitem[{Poliak et~al.(2018)Poliak, Naradowsky, Haldar, Rudinger, and
  Van~Durme}]{poliak-etal-2018-hypothesis}
Adam Poliak, Jason Naradowsky, Aparajita Haldar, Rachel Rudinger, and Benjamin
  Van~Durme. 2018.
\newblock \href {https://doi.org/10.18653/v1/S18-2023} {Hypothesis only
  baselines in natural language inference}.
\newblock In \emph{Proceedings of the Seventh Joint Conference on Lexical and
  Computational Semantics}, pages 180--191, New Orleans, Louisiana. Association
  for Computational Linguistics.

\bibitem[{Reimers and Gurevych(2019)}]{reimers-2019-sentence-bert}
Nils Reimers and Iryna Gurevych. 2019.
\newblock \href {http://arxiv.org/abs/1908.10084} {Sentence-bert: Sentence
  embeddings using siamese bert-networks}.
\newblock In \emph{Proceedings of the 2019 Conference on Empirical Methods in
  Natural Language Processing}. Association for Computational Linguistics.

\bibitem[{Ribeiro et~al.(2016)Ribeiro, Singh, and Guestrin}]{ribeiro2016should}
Marco~Tulio Ribeiro, Sameer Singh, and Carlos Guestrin. 2016.
\newblock " why should i trust you?" explaining the predictions of any
  classifier.
\newblock In \emph{Proceedings of the 22nd ACM SIGKDD international conference
  on knowledge discovery and data mining}, pages 1135--1144.

\bibitem[{Sadat and Caragea(2022)}]{sadat-caragea-2022-scinli}
Mobashir Sadat and Cornelia Caragea. 2022.
\newblock \href {https://doi.org/10.18653/v1/2022.acl-long.511} {{S}ci{NLI}: A
  corpus for natural language inference on scientific text}.
\newblock In \emph{Proceedings of the 60th Annual Meeting of the Association
  for Computational Linguistics (Volume 1: Long Papers)}, pages 7399--7409,
  Dublin, Ireland. Association for Computational Linguistics.

\bibitem[{Sanh et~al.(2019{\natexlab{a}})Sanh, Debut, Chaumond, and
  Wolf}]{DBLP:journals/corr/abs-1910-01108}
Victor Sanh, Lysandre Debut, Julien Chaumond, and Thomas Wolf.
  2019{\natexlab{a}}.
\newblock \href {http://arxiv.org/abs/1910.01108} {Distilbert, a distilled
  version of {BERT:} smaller, faster, cheaper and lighter}.
\newblock \emph{CoRR}, abs/1910.01108.

\bibitem[{Sanh et~al.(2019{\natexlab{b}})Sanh, Debut, Chaumond, and
  Wolf}]{Sanh2019DistilBERTAD}
Victor Sanh, Lysandre Debut, Julien Chaumond, and Thomas Wolf.
  2019{\natexlab{b}}.
\newblock Distilbert, a distilled version of bert: smaller, faster, cheaper and
  lighter.
\newblock \emph{ArXiv}, abs/1910.01108.

\bibitem[{Suntwal et~al.(2019)Suntwal, Paul, Sharp, and
  Surdeanu}]{suntwal-etal-2019-importance}
Sandeep Suntwal, Mithun Paul, Rebecca Sharp, and Mihai Surdeanu. 2019.
\newblock \href {https://doi.org/10.18653/v1/D19-1340} {On the importance of
  delexicalization for fact verification}.
\newblock In \emph{Proceedings of the 2019 Conference on Empirical Methods in
  Natural Language Processing and the 9th International Joint Conference on
  Natural Language Processing (EMNLP-IJCNLP)}, pages 3413--3418, Hong Kong,
  China. Association for Computational Linguistics.

\bibitem[{Wang et~al.(2020)Wang, Wei, Dong, Bao, Yang, and
  Zhou}]{wang2020minilm}
Wenhui Wang, Furu Wei, Li~Dong, Hangbo Bao, Nan Yang, and Ming Zhou. 2020.
\newblock \href {http://arxiv.org/abs/2002.10957} {Minilm: Deep self-attention
  distillation for task-agnostic compression of pre-trained transformers}.

\bibitem[{Wang et~al.(2019)Wang, Wang, Duan, Wu, Wang, Hu, and
  Liu}]{wang-etal-2019-iflylegal}
Ziyue Wang, Baoxin Wang, Xingyi Duan, Dayong Wu, Shijin Wang, Guoping Hu, and
  Ting Liu. 2019.
\newblock \href {https://doi.org/10.18653/v1/D19-3017} {{IF}ly{L}egal: A
  {C}hinese legal system for consultation, law searching, and document
  analysis}.
\newblock In \emph{Proceedings of the 2019 Conference on Empirical Methods in
  Natural Language Processing and the 9th International Joint Conference on
  Natural Language Processing (EMNLP-IJCNLP): System Demonstrations}, pages
  97--102, Hong Kong, China. Association for Computational Linguistics.

\bibitem[{Wasserblat(2021)}]{wasserblat_2021}
Moshe Wasserblat. 2021.
\newblock \href
  {https://towardsdatascience.com/sentence-transformer-fine-tuning-setfit-outperforms-gpt-3-on-few-shot-text-classification-while-d9a3788f0b4e}
  {Sentence transformer fine-tuning (setfit): Outperforming gpt-3 on few-shot
  text-classification while being 1600 times smaller}.

\end{thebibliography}
\bibliographystyle{acl_natbib}

\end{document}